\documentclass[runningheads]{llncs}
\usepackage[T1]{fontenc}
\usepackage{amsmath,amsfonts}
\usepackage{subcaption}
\usepackage{multirow}   
\usepackage{booktabs}   
\usepackage{graphicx}
\usepackage{svg}
\usepackage{enumitem}
\usepackage{algorithm}
\usepackage{algpseudocode}
\usepackage{booktabs} 
\usepackage{breakurl}

\usepackage{graphicx}

\usepackage{cite}
\usepackage{multiobjective}
\usepackage{todonotes}  %  use [disable] to... well... disable
\setuptodonotes{size=tiny,color=pink!29}
\newcommand{\errCESUniOne}{$4.4 \times 10^{-6}~(\pm~4.2 \times 10^{-6})$}
\newcommand{\srCESUniOne}{100\%}
\newcommand{\errCBOUniOne}{$2.9 \times 10^{-5}~(\pm~2.2 \times 10^{-5})$}
\newcommand{\srCBOUniOne}{100\%}
\newcommand{\errSGDUniOne}{$0.04~(\pm~9.3 \times 10^{-18})$}
\newcommand{\srSGDUniOne}{0\%}
\newcommand{\errCESUniTwo}{$7.4 \times 10^{-4}~(\pm~3.9 \times 10^{-4})$}
\newcommand{\srCESUniTwo}{100\%}
\newcommand{\errCBOUniTwo}{$1.3 \times 10^{-4}~(\pm~6.9 \times 10^{-5})$}
\newcommand{\srCBOUniTwo}{100\%}
\newcommand{\errSGDUniTwo}{$0.02~(\pm~4.7 \times 10^{-18})$}
\newcommand{\srSGDUniTwo}{0\%}
\newcommand{\errCESUniTen}{$0.14~(\pm~0.10)$}
\newcommand{\srCESUniTen}{0\%}
\newcommand{\errCBOUniTen}{$0.02~(\pm~5.7 \times 10^{-3})$}
\newcommand{\srCBOUniTen}{0\%}
\newcommand{\errCESUniThirty}{$3.51~(\pm~0.29)$}
\newcommand{\srCESUniThirty}{0\%}
\newcommand{\errCBOUniThirty}{$0.26~(\pm~0.09)$}
\newcommand{\srCBOUniThirty}{0\%}
\newcommand{\errCESShiOne}{$4.9 \times 10^{-6}~(\pm~3.8 \times 10^{-6})$}
\newcommand{\srCESShiOne}{100\%}
\newcommand{\errCBOShiOne}{$0.08~(\pm~0.06)$}
\newcommand{\srCBOShiOne}{0\%}
\newcommand{\errSGDShiOne}{$1.97~(\pm~0)$}
\newcommand{\srSGDShiOne}{0\%}
\newcommand{\errCESShiTwo}{$0.49~(\pm~0.54)$}
\newcommand{\srCESShiTwo}{43\%}
\newcommand{\errCBOShiTwo}{$1.29~(\pm~0.20)$}
\newcommand{\srCBOShiTwo}{0\%}
\newcommand{\errSGDShiTwo}{$2.79~(\pm~0)$}
\newcommand{\srSGDShiTwo}{0\%}
\newcommand{\errCESShiTen}{$3.86~(\pm~0.40)$}
\newcommand{\srCESShiTen}{0\%}
\newcommand{\errCBOShiTen}{$5.19~(\pm~0.32)$}
\newcommand{\srCBOShiTen}{0\%}
\newcommand{\errCESShiThirty}{$8.74~(\pm~0.38)$}
\newcommand{\srCESShiThirty}{0\%}
\newcommand{\errCBOShiThirty}{$10.20~(\pm~0.25)$}
\newcommand{\srCBOShiThirty}{0\%}
\begin{document}
\title{From Mean-Field Limits to Semiclassical Concentration: Global Convergence of the Canonical Evolutionary Strategy}
\titlerunning{From Mean-Field Limits to Semiclassical Concentration: Convergence of CES}
%% BLIND SUBMISSION
\author{Matías~Neto\inst{1}\and
Nicolás~Garay\inst{2}\thanks{Work done during internship at Inria Chile Research Center.}\and
Luis~Martí\inst{1} \and
Nayat~Sanchez-Pi\inst{1}}
\authorrunning{M. Neto et al.}
\institute{Inria Chile Research Center, Av Apoquindo 2827, Las Condes, RM, Chile  \email{\{matias.neto, luis.marti, nayat.sanchez-pi\}@inria.cl}, \and
Departamento de Ingeniería Matemática, Universidad de Chile, Santiago, Chile
\email{ngaray@dim.uchile.cl}}
\maketitle              % typeset the header of the contribution
%\todo[inline]{Reviewer -- global editorial issues to address before submission:
%(1) Duplicate labels: build log reports Label fig:branching\_validation\_\_b multiply defined --- please search for repeated \textbackslash{}label and fix. (2) Overfull hboxes on lines 66--67, 90--93, 303--304. (3) Underfull hboxes in the bibliography (long URLs) --- consider breakurl. (4) SVG warning: Inkscape not found; convert SVGs to PDF at source. (5) Typos: ``reproductibility'' $\to$ ``reproducibility'' (\S4.1); ``slighty'' $\to$ ``slightly'' (\S4.2); ``CBO reach'' $\to$ ``CBO reaches'' (Fig.~5 caption); Fig.~3 caption repeats $(\sigma=0.04)$ twice. (6) Fig.~5 y-axis labels appear truncated (``$8\times 10$'' etc). (7) Consider adding a notation table after \S2; the sign switch between $g$ (max) and $\ell$ (min, Remark~1) is easy to miss.}
%
\begin{abstract}
%\todo[inline]{Reviewer: The abstract and Introduction overlap substantially (several sentences are near-verbatim). Consider tightening one or the other. Also, ``global minimizer'' vs ``global maximum'' is used inconsistently across the paper; please fix the sign convention once for the entire manuscript.}
%\todo[inline]{Reviewer: Please state explicitly here that the \emph{theoretical} results (mean-field + semiclassical concentration) are proved only for $d=1$ on a bounded $\Omega\subset\mathbb{R}$, while the \emph{empirical} claims extend to $d=30$ on $\mathbb{R}^d$. Readers risk mistaking the d=1 theorem for a general guarantee.}
We address the issue of global convergence in stochastic continuous optimization. For that purpose, we formulate the Canonical Evolutionary Strategy (CES) as a controlled mathematical framework to analyze global convergence in evolutionary algorithms via the semiclassical limit of a Schrödinger-type replicator-mutator equation. We provide a rigorous hierarchy from a discrete individual-based dynamics to a deterministic mean-field limit, demonstrating that global convergence is governed by the principal eigenfunction of the underlying operator. This property, defined as Geometric Selection, naturally prioritizes robust, flat optima over narrow local traps, offering a mathematical justification for the ``survival of the flattest'' phenomenon. Moreover, unlike consensus-driven methods that are prone to premature variance collapse when the global minimizer resides outside the initial support, the replicator-mutator dynamics of CES facilitate intrinsic mass transport. High-dimensional benchmarks ($d=30$) confirm this advantage, showing that CES achieves lower residual errors in shifted initialization scenarios where standard consensus-driven and gradient-based methods fail to migrate effectively. By shifting the focus from point-wise consensus to spectral concentration, our framework provides a robust theoretical foundation for global convergence in Evolution Strategies (ES) without the need for additional numerical heuristics.

\end{abstract}
%
%\listoftodos%
%
\section{Introduction}
Global convergence in stochastic optimization remains a fundamental challenge in computational mathematics. While metaheuristic approaches -- such as genetic algorithms and evolutionary strategies (ES) \cite{hansen2001completely, gendreau2013handbook} -- have shown immense empirical success in navigating non-convex landscapes, they often lack rigorous theoretical guarantees, particularly regarding global convergence to the optimum in high-dimensional settings \cite{roith2025consensus}. Establishing methods that can provably bypass local entrapment without depending on local derivative information is essential for solving complex problems in machine learning, robotics, and scientific computing \cite{williams2017mode,herty2024multiscale,fornasier2026}.

A key development in this direction is Consensus-Based Optimization (CBO) \cite{Pinnau2016ConsensusMeanField}, a multi-agent framework designed to solve global optimization problems with provable convergence. CBO relies on a balance between stochastic exploration and a collective contraction toward a consensus point, which acts as a proxy for the global minimizer. However, the convergence of CBO is often tied to the requirement that the global minimizer resides within the initial support of the individual distribution. When this condition is not met, the system is prone to \emph{premature collapse}, where the population variance vanishes into a sub-optimal state before effective migration can occur \cite{fornasier2026}.

One way of approaching this issue is again turn to nature for inspiration, and, in particular to evolutionary ecology \cite{10.1098/rsfs.2013.0054} results that rely on the mean-field theory \cite{refId0}. 

With that goal in mind, we formulate the Canonical Evolutionary Strategy (CES) to establish a rigorous framework for global convergence in evolutionary dynamics. Inspired by the probabilistic limits developed in evolutionary ecology \cite{fournier2004microscopic, champagnat2006unifying}, we derive this macroscopic transition using the semiclassical limit of a Schrödinger-type replicator-mutator equation \cite{pathiraja2024mathematicaldescriptioncontinuoustime}. This perspective shifts the focus from point-wise consensus to global mass transport, providing a formal guarantee that the population density concentrates on the global optimum, regardless of its initial support.

The central contribution of this paper is that the interplay between selection and diffusion in the CES provides a more robust mechanism for global convergence than traditional consensus-driven models. We establish a rigorous hierarchy from discrete individual-based dynamics to a deterministic mean-field limit, demonstrating that in the semiclassical limit, the population mass is mathematically compelled to concentrate on the global maximum via the principal eigenfunction of the system. This \emph{geometric selection} naturally prioritizes robust, flat optima over narrow local traps. We validate this theoretical foundation with high-dimensional benchmarks ($d=30$), demonstrating that CES consistently achieves lower residual errors and deeper convergence in \textit{shifted initialization} scenarios where standard consensus-based and gradient-descent methods fail to migrate effectively.

The remainder of the paper is organized as follows. Section~2 introduces the CES dynamics and specifies the discrete stochastic model studied throughout the paper. Section~3 develops the theoretical framework, deriving the mean-field limit and presenting the semiclassical perspective that links global convergence to the principal eigenfunction of the associated Schrödinger-type operator. Section~4 studies the stationary regime and the resulting concentration mechanism in the one-dimensional setting, while Section~5 complements the analysis with numerical experiments, including shifted-initialization benchmarks in dimension $d=30$ and comparisons against CBO-based baselines. Finally, Section~6 summarizes the main conclusions and discusses directions for extending the framework to higher-dimensional settings and richer evolutionary operators.

\section{The Canonical Evolutionary Strategy (CES)}
%\todo[inline]{Reviewer: The label ``Canonical ES'' may be misleading to the EC community, which typically reserves ``canonical/standard ES'' for $(\mu,\lambda)$- or $(\mu/\rho,\lambda)$-ES with rank-based (deterministic truncation) selection and self-adaptive step sizes (Hansen \& Ostermeier, Beyer \& Schwefel). The scheme analysed here uses fitness-proportional selection and a fixed additive Gaussian perturbation, which is closer to a Wright--Fisher / replicator--mutator model than to what practitioners call an ES. Please either (a) rename the algorithm (e.g. ``Fitness-Proportional ES'' or ``Replicator--Mutator ES'') or (b) add an explicit remark here acknowledging the terminological distinction and positioning the work relative to CMA-ES and related canonical ES variants.}
We now define the stochastic optimization algorithm analyzed throughout this work. We consider the global optimization problem of finding the maximizer of a non-negative fitness function $g:\mathbb{R}^d\to \mathbb{R}_+$
\begin{equation}
    \argmax_{x\in \mathbb{R}^d} g(x)
\end{equation}
The Canonical Evolutionary Strategy (CES)\footnote{Not to be confused with the standard $(\mu,\lambda)$- or $(\mu/\rho,\lambda)$-ES.} is modeled as a discrete-time Markov chain $(Y^M (n))_{n\in \mathbb{N}}$ representing a population of $M$ individuals in a continuous search space. Each iteration $n$ of the algorithm consists of the following three stages
\begin{enumerate}
    \item \emph{Initialization}: At $n=0$, an initial population $Y^M(0)=\left(Y_1^{M}(0),\ldots,Y_M^{M}(0)\right)$ is generated. Each individual $Y^M_j(0)$ is sampled independently from an initial probability density function $f_0$.
    \item \emph{Selection}: Given the population at step $n$, a set of $M$ intermediate individuals $\widetilde{Y}^{M}(n)=\left(\widetilde{Y}_1^{M}(n),\ldots,\widetilde{Y}_M^{M}(n)\right)$ is selected. The selection is performed with replacement, where the probability of choosing an individual $Y^M_j(n)$ is proportional to its fitness value
    \begin{equation}
        \mathbb{P}\!\left(\widetilde{Y}_i^{M}(n)=Y_j^{M}(n)\,\middle|\,Y^{M}(n)\right)
        =\frac{g\!\left(Y_j^{M}(n)\right)}{\sum_{k=1}^{M} g\!\left(Y_k^{M}(n)\right)},
        \ j\in\{1,\ldots,M\}.
    \end{equation}
    \item \emph{Variation}: The next generation $Y^M(n+1)$ is produced by applying an additive stochastic perturbation to the selected individuals
    \begin{equation}
        Y_i^{M}(n+1)\sim \widetilde{Y}_i^{M}(n)+f_n,\qquad i\in\{1,\ldots,M\}.
    \end{equation}
    where $(f_n)$ are known probability densities, typically chosen as zero-mean Gaussian distributions to explore the neighborhood of selected candidates.
    \end{enumerate}
This formulation allows us to track the empirical distribution of the population as it evolves. In the following sections, we justify how an appropriate scaling of the fitness ($g$) and mutation ($f_n$) relative to the population size ($M$) allows this discrete process to converges to a continuous-time mean-field limit. This limit allows us to characterize the algorithm's global convergence properties through a semiclassical analysis of the underlying operator.

\section{Theoretical Framework: A Mean-Field Limit Approach}

The following results establish a bridge between the discrete stochastic population and the deterministic search for global optima. While the formal mean-field derivation presented here focuses on the one-dimensional case ($d=1$), the resulting semiclassical framework provides the fundamental intuition for the algorithm's behavior on higher dimensions.
%\todo[inline]{Reviewer: This one sentence is doing a lot of work. Please expand into a short paragraph that (i) states which theorems are for $d=1$ only (Theorem~\ref{th:semi_class} and the mean-field limit), (ii) identifies the technical obstacles to $d>1$ (compactness of $\Omega$, spectral gap, boundary conditions, ground-state uniqueness on unbounded domains), and (iii) clearly labels the $d>1$ claims as a \emph{conjecture} supported by numerics. Without this, the $d=30$ experimental claims may appear to rest on a theorem that does not exist.}

\subsection{Mean-Field Limit: From Individuals to Partial Differential Equations (PDEs)}

The first fundamental step is to establish a connection between the stochastic CES and a deterministic continuous-time model. Such large-population approximations from microscopic individual-based models to macroscopic PDEs are well-established in the study of adaptive dynamics \cite{fontbona2015nonlocal,meleard2016modeles}. To achieve this, we adopt the scaling limit proposed by Wakano \emph{et\,al.} \cite{wakano2017derivation}. We consider a search space $\Omega\subset \mathbb{R}$ (typically a bounded interval) and a measurable bounded fitness function $\sigma \in B(\Omega)$.

For a population size $M$, we define the selection intensity as $a_M$ and the mutation time step as $t_M$. To guarantee convergence as $M\to \infty$, we assume $a_M\to 0$, $\tfrac{a_M}{t_M}\to a$, and $t_M M\to \infty $. In our practical framework, we follow the power-law scaling
\begin{equation}
    a_M := a M^{-\alpha}, t_M:=M^{-\alpha}, ~ a>0, ~ \alpha \in (0,1).
\end{equation}
Under these conditions, the CES stages are formally redefined as follows
\begin{enumerate}
    \item \emph{Selection:} Intermediate individuals $\widetilde{Y}_i^{M}(n)$ are sampled with weights 
    \begin{equation}
        w_M(x):=1+a_M\sigma(x).
    \end{equation}
    \item \emph{Variation:} The next generation is produced using the heat kernel $p(c\, t_M,\cdot,\cdot)$ on $\Omega$ with Neumann boundary conditions
    \begin{equation}
        Y_i^{M}(n+1)\sim p(c\, t_M, \cdot, \widetilde{Y}_i^{M}(n)),\quad i\in\{1,\ldots,M\}.
    \end{equation}
\end{enumerate}

The transition from an individual system to a continuous density is captured by the following theorem:

\begin{theorem}{Mean Field-Limit for CES \cite[Theorem 2.1]{wakano2017derivation}.}\label{th:mean-field}
   Let $\nu^M(t)$ be the continuous-time process defined by the empirical measure of the population at the scaled steps $\lfloor t_M^{-1}t \rfloor$. If the initial population $\nu^M(0)$ converges in law to a density function $f_0$, then $\nu^M(t)$ converges in law to the unique solution of the replicator-mutator equation with Neumann boundary conditions
   \begin{equation}\label{eq:rep_mut}
   \begin{cases}
    \partial_t u = c\Delta u + a\bigl(\sigma(x)-\sigma(u)\bigr)\,u,
    & t>0,\ x\in \Omega,\\
    \partial_n u = 0,
    & t>0,\ x\in \partial \Omega,\\
    u(0) = f_0,
    & x\in \Omega.
    \end{cases}
    \end{equation}
    where $\sigma(u)=\int_\Omega \sigma(x)u(t,x)dx$, $a$ and $c$ are the mutation and selection coefficients, respectively.
\end{theorem}

It implies that for a sufficiently large population $M$, the evolution of the stochastic algorithm is effectively governed by the dynamics of the replicator-mutator PDE. Although it may seem counter intuitive that selection and mutation intensities vanish at each step ($a_M,t_M\to 0$), this scaling is precisely what allows the accumulation of many infinitesimal changes to yield a stable continuous trajectory. By increasing the total number of iterations $\lfloor t_M^{-1}t \rfloor$ as $M$ grows, we compensate for the reduced intensity per step, ensuring that the macroscopic evolutionary signal remains consistent and predictable.

\subsubsection{Numerical Validation}

To empirically validate Theorem \ref{th:mean-field}, we consider a one-dimensional domain $\Omega=\left[-2,2\right]$ and a fitness function $\sigma_1(x)$ characterized by a single global maximum and a local trap
\begin{equation}
\sigma_1(x) = 0.2 e^{-5x^4 + x^3 + x^2 + 1}.
\end{equation}
A key practical challenge is that the mean-field limit formally requires a total of $\lfloor T M^\alpha \rfloor$ iterations. To ensure computational efficiency without losing the concentration properties, we employ a temporal scaling argument: if $u(t,x)$ is a solution with parameters $a$ and $c$, then $\widetilde{u}(t,x):=u(Tt,x)$ satisfies the same PDE but with accelerated coefficients $\widetilde{a}=aT$ and $\widetilde{c}=cT$. For our simulations, we choose $a=40$, $c=0.04$ with $M=2\cdot 10^5$, $\alpha=0.5$ and $T=3.0$. This allows the population to reach its stationary state in only $\lfloor T M^\alpha \rfloor \approx 1341$ iterations.

We compare the stochastic trajectory of the population against the numerical solution of the replicator-mutator PDE \eqref{eq:rep_mut}, which we solved using a Crank-Nicolson scheme for the diffusion term and an explicit scheme for the selection term. To quantify the discrepancy between the empirical measure $\nu^M(t)$ and the continuous density $u(t,\cdot)$, we employ the $L^1$-Wasserstein distance 
\begin{equation}
    W_1(u(t,\cdot),\nu^M(t))=\int_\Omega |F_{u(t,x)}-F_{\nu^M(t,x)}|dx,
\end{equation}
where $F$ denotes the cumulative distribution function. Now we can properly define the time-averaged $L^1$-Wasserstein distance
\begin{equation}
    \bar{W_1}=\frac{1}{T}\int^{T}_0 W_1(u(t,\cdot),\nu^M(t)) dt.
\end{equation}
This metric provides a global measure of how closely the stochastic individual system tracks the theoretical PDE trajectory throughout the entire optimization process.
\begin{figure}[tb]
\centering
\begin{subfigure}{0.48\textwidth}
\centering
\includegraphics[width=\linewidth]{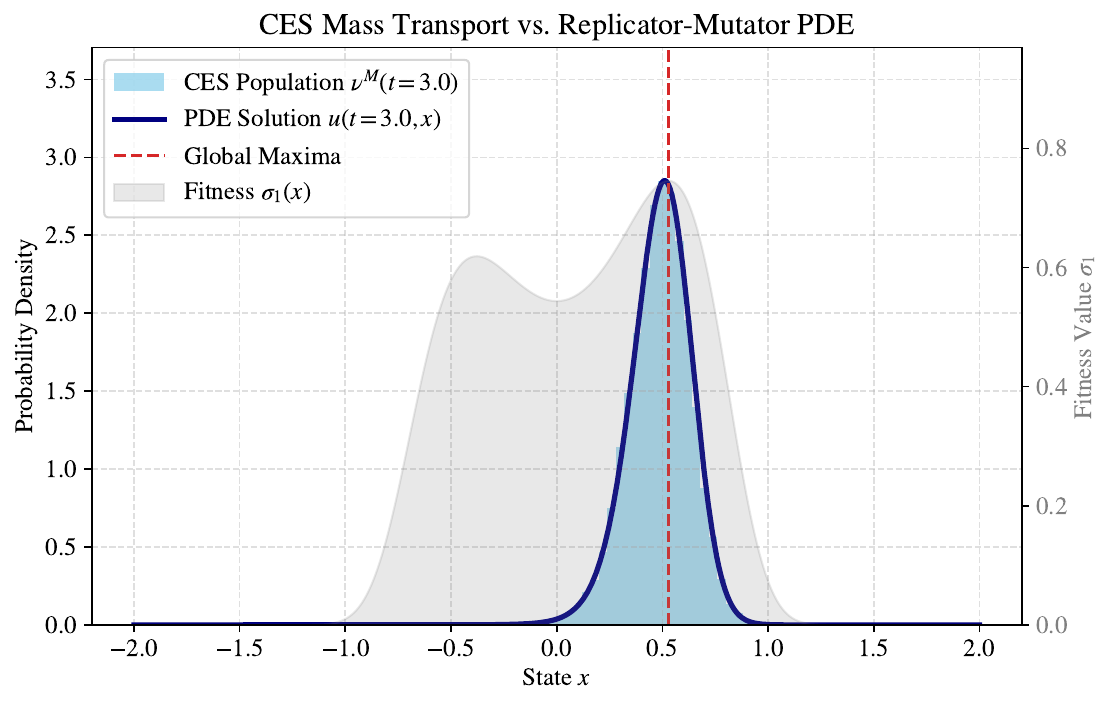} 
\caption{Evolution of $\nu^M(t)$ ($M=2\cdot 10^5,~a=40,~c=0.04$).}\label{fig:validation_mean_field__evol}
\end{subfigure}
\hfill
\begin{subfigure}{0.48\textwidth}
\centering
\includegraphics[width=\linewidth]{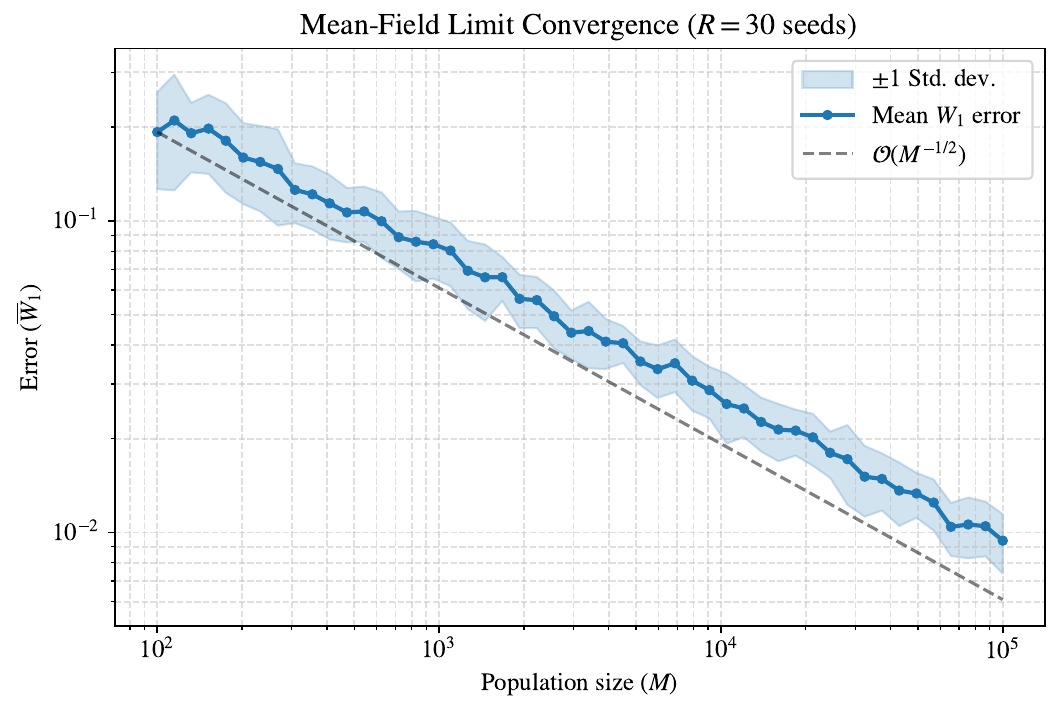}
\caption{Time averaged $W_1$ error vs. Population Size $M$ ($T=3$).}\label{fig:validation_mean_field__timeavg}
\end{subfigure}
\caption{\textbf{Mean-Field Limit Validation}. (a): The population distribution $\nu^M(t)$ follows the PDE dynamics, escaping the local trap. (b): The time-averaged Wasserstein distance between the stochastic CES and the replicator-mutator PDE as a function of population size M. The solid line denotes the mean error over $30$ independent seeds, while the shaded area represents the $\pm1$ standard deviation. The log-log scale highlights the consistent convergence rate toward the continuous limit.}
\label{fig:validation_mean_field}
\end{figure}

As observed in Fig.~\ref{fig:validation_mean_field__evol}, the population effectively concentrates around the global maximum. The convergence rates observed in Fig.~\ref{fig:validation_mean_field__timeavg} align with the expected $\mathcal{O}(M^{-\tfrac{1}{2}})$ decay rate for empirical measure approximations of interacting particle systems \cite{fornasier2026}. As $M$ increases, the stochastic fluctuations of the CES empirical measure $\nu^M$ vanish, effectively recovering the deterministic behavior of the mean-field limit $u$.

\subsection{Asymptotic Behavior: Convergence to global maxima}

While the mean-field limit describes the trajectory of the population, establishing global optimization requires ensuring that the CES settles into a stable equilibrium. We analyze the long-term behavior of the replicator-mutation equation ($t\to \infty$) by relating it to the spectral properties of an associated Schrödinger-type operator.

The stationary solution of the PDE is determined by the principal eigenpair of the operator $L=-c\Delta -a\sigma$ with Neumann boundary conditions. We state the following convergence property
\begin{theorem}{Asymptotic convergence \cite[Theorem 1.2]{Coville2013Convergence}.}\label{th:cv_prop} Let $u(t,x)$ be a positive solution to the replicator-mutation equation \eqref{eq:rep_mut} in a bounded Lipschitz domain $\Omega$. Let $\lambda_1$ be the principal eigenvalue and $\phi_1$ the corresponding positive principal eigenfunction of $L$, normalized in $L^2(\Omega)$, satisfying
\begin{equation}\label{eq:eigen_vec}
\begin{cases}
-c\Delta \phi_1 - a\sigma\phi_1 = \lambda_1 \phi_1, & x \in \Omega \\
\partial_n \phi_1 = 0, & x \in \partial \Omega
\end{cases}
\end{equation}
Then, the asymptotic behavior of the population density $u(t,x)$ depends on the sign of $\lambda_1$
\begin{enumerate}
\item If $\lambda_1\ge0$, there are no positive stationary solutions, and $u(t,x)\to 0$ as $t\to \infty$.
\item If $\lambda_1<0$, the population converges to the unique stationary density
\begin{equation}
    u(t,x)\longrightarrow \mu \phi_1(x)\, \text{ as }t\to\infty
\end{equation}
where the normalization constant is $\mu:=\left(\int_\Omega \phi_1(y) dy \right)^{-1}$.
\end{enumerate}
\end{theorem}
Establishing the long-term behavior of the population is fundamental for global optimization. While the specific value of $\lambda_1$ depends on the complex interplay between the selection pressure ($a$), the mutational diffusion ($c$) and the fitness landscape ($\sigma$), the case $\lambda_1 <0$ is the one of interest as it ensures the existence of a unique stationary density $\phi_1$, \emph{i.e.}, ensures that the evolutionary process is asymptotically stable.

However, for optimization purposes, the mere existence of an equilibrium is insufficient; we require this density to concentrate its mass exclusively on the global maximizers of $\sigma$. To address this, we treat the ratio between mutation and selection as a semiclassical parameter $h:=\tfrac{c}{a}$. As $h\to 0$ (which corresponds to high selection pressure relative to mutation), we can apply the tools of semiclassical analysis to study the concentration of the population mass.

The stationary density $\phi_1 (h)$ satisfies the following semiclassical Schrödinger equation
\begin{equation}\label{eq:eigen_vec_2}
\begin{cases}
-h^2\Delta \phi_1 -\sigma\phi_1\;=\; \widetilde{\lambda_1} \phi_1& x\in \Omega\\[4pt]
\partial_n \phi_1 \;=\; 0,
&  x\in \partial \Omega
\end{cases}
\end{equation}
where $\widetilde{\lambda_1}(h):=\tfrac{\lambda_1}{a}$.

The following property is a direct consequence of the variational characterization of the principal eigenvalue \cite[Section 2.2]{cantrell2003spatial} evaluated in the semiclassical limit \cite{Zworski2012Semiclassical}. It provides the mathematical guarantee for the CES.

\begin{theorem}{Semiclassical Concentration \cite{cantrell2003spatial, Zworski2012Semiclassical}.}\label{th:semi_class}
    Assume $\sigma\in \mathcal{C}(\bar{\Omega})$ has at least one global maximum in $\Omega$. Then
    \begin{enumerate}
        \item $\lim_{h\to 0} \widetilde{\lambda_1}(h)=-\max_{x\in \Omega}\sigma(x)$. This ensures that for a small enough $h$, $\lambda_1<0$, satisfying the stability condition of Theorem \ref{th:cv_prop}.
        \item For any compact set $U\subset \Omega$ that does not contain a global maximizer of $\sigma$
        \begin{equation}
        \lim_{h \to 0} \frac{|\phi_1(h)|_{L^2(U)}}{|\phi_1(h)|_{L^2(\Omega)}} = 0.
        \end{equation}
    \end{enumerate}
\end{theorem}

\subsubsection{Numerical Validation}

\begin{figure}[tb]
\centering
\includegraphics[width=0.6\linewidth]{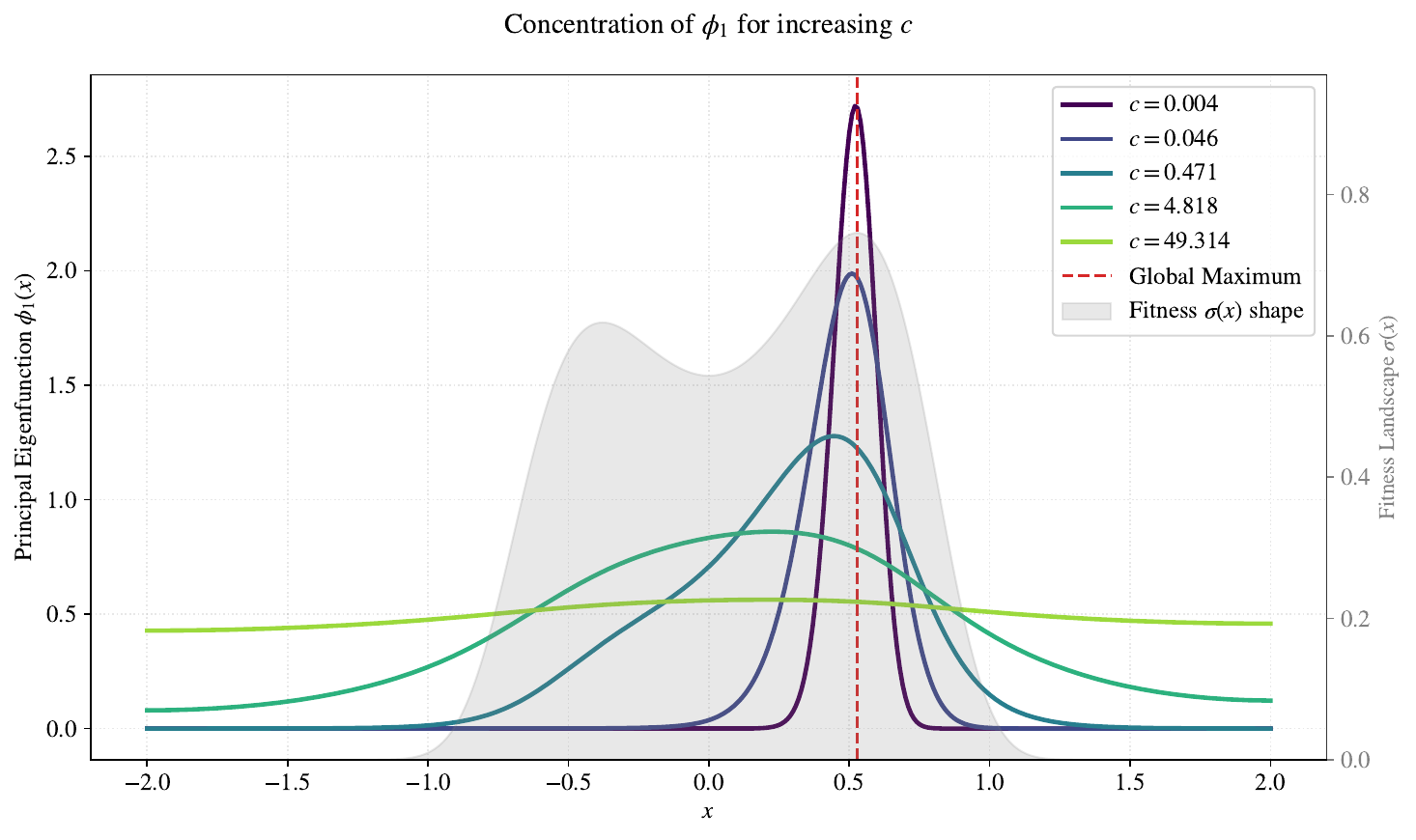}
\caption{\textbf{Semiclassical Concentration ($\sigma_1$)}. Evolution of the stationary density $\phi_1$ for variable $c$ with fixed $a=40$. The mass progressively concentrates at the global maximum, effectively bypassing local trap as $h\to 0$.}
\label{fig:semiclassical_limit}
\end{figure}

To characterize the semiclassical limit, we analyze the concentration of the stationary distribution $\phi_1$ in the $\sigma_1$ landscape. As shown in Figure \ref{fig:semiclassical_limit}, by reducing the mutational diffusion $c>0$ (which corresponds to $h\to 0$), the population mass localizes around the global maximizer. This numerical behavior confirms that the ratio $h^2=\tfrac{c}{a}$ dictates the resolution of the global search.

\begin{figure}[tb]
\centering
\begin{subfigure}{0.48\textwidth}
\centering
\includegraphics[width=\textwidth]{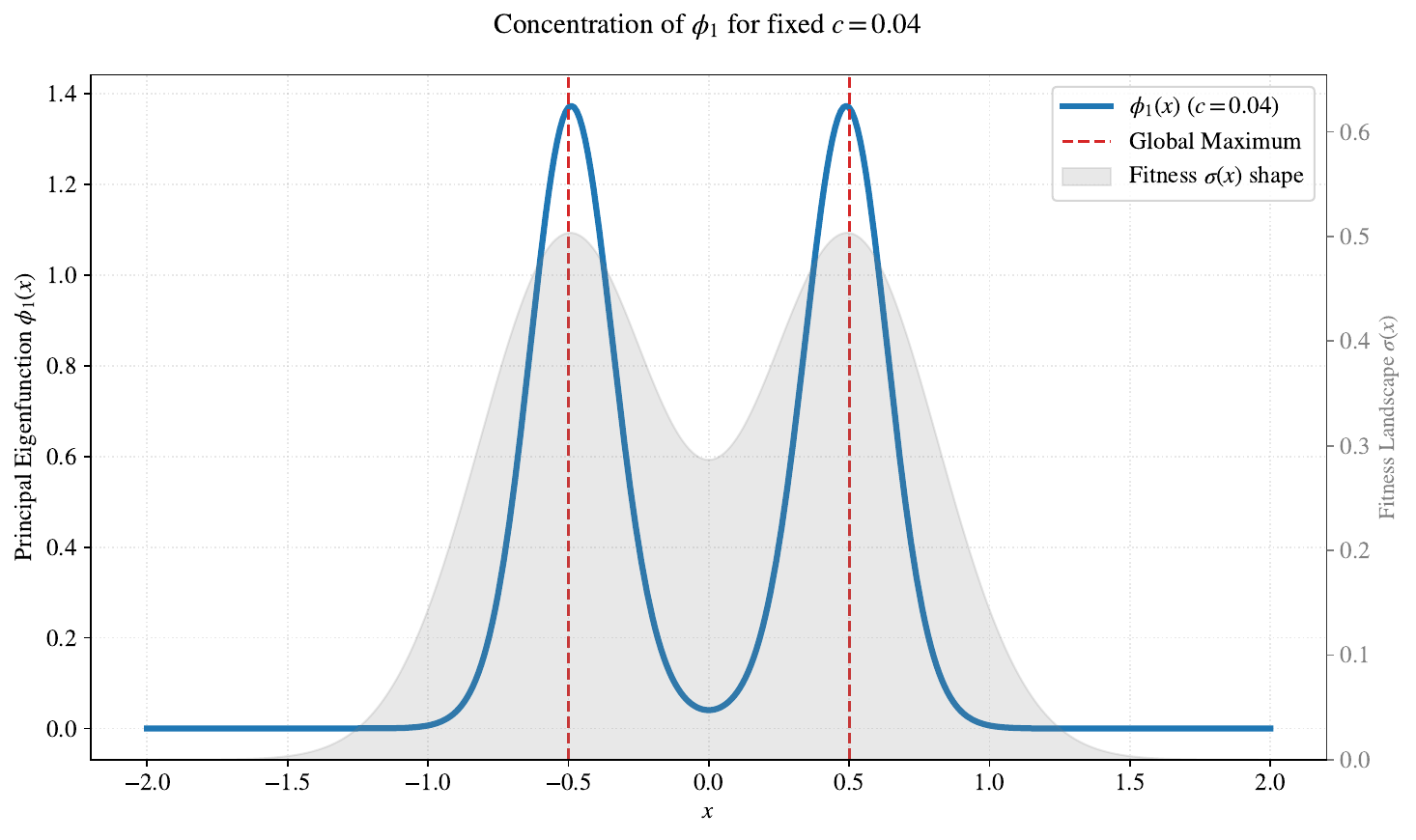}
\caption{Experiment 2: $\sigma_2$.}\label{fig:branching_validation__a}
\end{subfigure}
\hfill
\begin{subfigure}{0.48\textwidth}
\centering
\includegraphics[width=\textwidth]{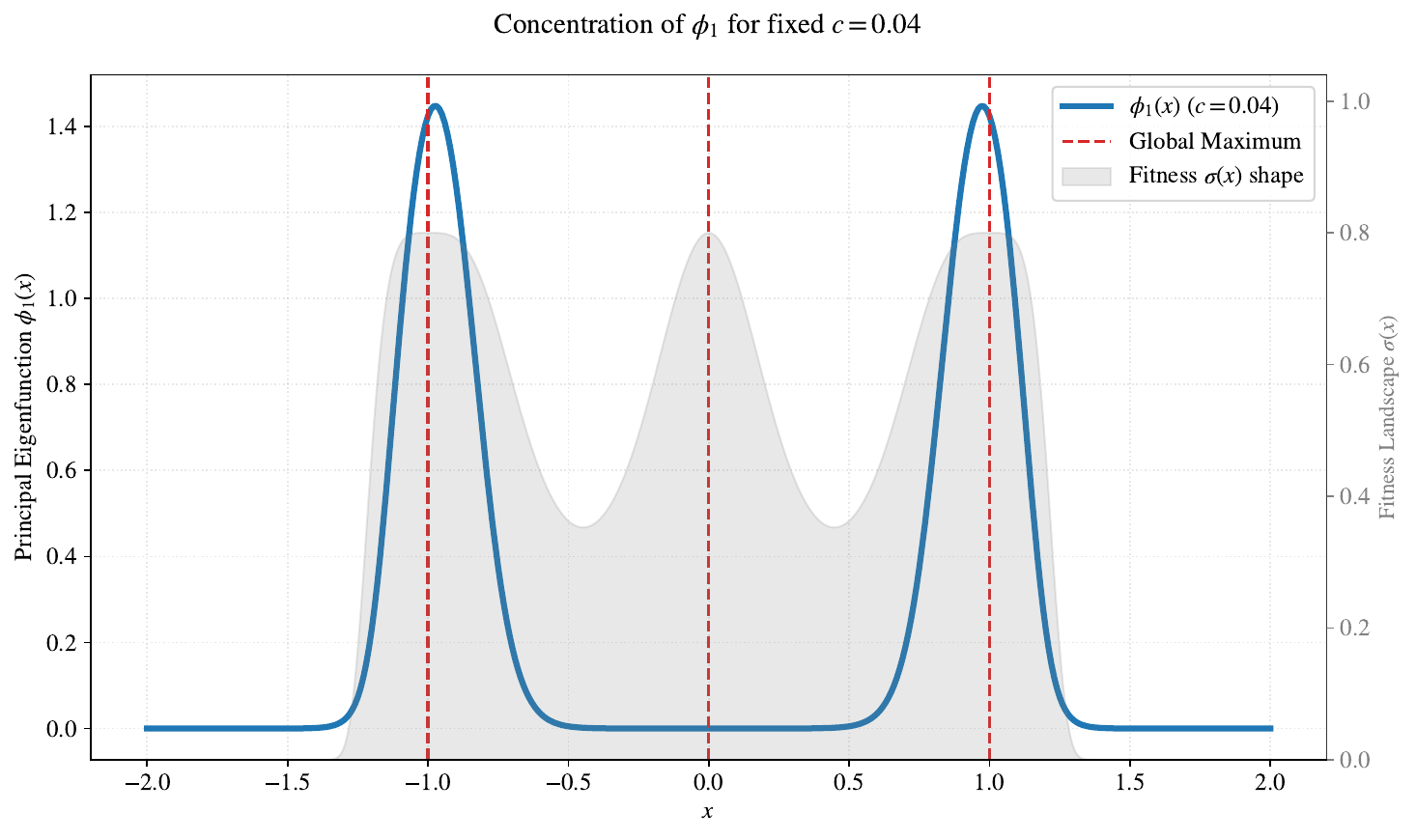}
\caption{Experiment 3: $\sigma_3$.}\label{fig:branching_validation__b}
\end{subfigure}
\caption{Stationary distributions in complex landscapes ($a=40, c=0.04$). (a) Characterization of the invariant measure in symmetric peaks. (b) Preference for robust (wider) optima under constant mutational pressure.}
\label{fig:branching_validation}
\end{figure}

The geometric properties of this equilibrium are further examined in multi-modal landscapes ($a=40$, $c=0.04$). Fig.~\ref{fig:branching_validation} illustrates two fundamental behaviors of the steady state $\phi_1$:
\begin{itemize}
    \item \emph{Symmetric branching ($\sigma_2$)}: In the presence of two identical global maxima, 
    $\sigma_2(x)=0.5\left[\exp\left({-5(x-5)^2}\right)+\exp\left({-5(x+5)^2}\right)\right]$, the population reaches a stable invariant measure that represents the optimal set $\mathcal{M}$ through a balanced mass split (see Fig.~\ref{fig:branching_validation}a).
    \item \emph{Geometric Selection ($\sigma_3$)}: In landscapes with global optima of varying width,
    $\sigma_3(x)=0.8\exp\left({-10x^2(x-1)^4(x+1)^4}\right)$, the population mass concentrates on the wider optima (see Fig.~\ref{fig:branching_validation}b). While Theorem \ref{th:semi_class} ensures concentration on the set of global maximizers as $h\to 0$, the stationary state prioritizes maximizers with lower mutational loss. This ``survival of the flattest'' effect, consistent with the $h^2\Delta$ term in our Schrödinger-type framework, suggests a resilience against narrow local traps - a property that potentially underpins the algorithm's robust performance in high-dimensional rugged landscapes.
\end{itemize}

\section{Experimental Results and Benchmarks}
%\todo[inline]{Reviewer: The empirical evaluation is too narrow for PPSN. Please broaden the benchmark in at least three directions:
%(1) \textbf{Test functions:} evaluate on a representative subset of BBOB/COCO (Rastrigin, Schwefel, Rosenbrock, Griewank, Weierstrass, and an ill-conditioned function such as Discus or BentCigar) to cover separable, ill-conditioned, multimodal with global structure, and multimodal with weak structure classes.
%(2) \textbf{Baselines:} add CMA-ES (the de facto ES baseline) and a modern CBO variant such as $\lambda$-CBO or the Consensus Freezing scheme of [7], since §5.2 explicitly compares against these qualitatively.
%(3) \textbf{Budget parity:} state the number of function evaluations per run explicitly and equalise it across methods; report whether the population size $M$ was tuned per method or held fixed.}
%\todo[inline]{Reviewer: Please also state the exact number of seeds used for Table~1 (the current phrasing says ``$N$ independent realization'' but $N$ appears to be a placeholder). Report either 95\% confidence intervals or a non-parametric test (Mann--Whitney / Wilcoxon) on the per-seed errors; the gap in Success Rate between 43\% (CBO) and 100\% (CES) in the Shifted $d=2$ case deserves an explicit significance statement.}
%\todo[inline]{Reviewer: Add a hyperparameter sensitivity study over the scaling exponent $\alpha\in(0,1)$ (Eq.~(4)), the population size $M$, and the mutation variance. Without this, it is hard to tell whether the CES advantage reflects the theoretical mechanism or favourable tuning.}

In this section, we evaluate the performance and robustness of the proposed CES. Our analysis focuses on the transition from local exploitation to global mass transport, comparing the CES framework against established optimization paradigms across different dimensionalities and initialization conditions.

\subsection{Experimental Setup}

To ensure statistical significance and reproducibility, all reported results are based on $30$ independent realization with distinct random seeds. We provide a rigorous comparison between the following methods:
\begin{itemize}
    \item \textbf{CES:} The proposed strategy based on the semiclassical concentration of the Schrödinger-type invariant measure.
    \item \textbf{Consensus-Based Optimization (CBO) \cite{Pinnau2016ConsensusMeanField}:} A state-of-the-art framework for global optimization derived from mean-field dynamics. We select CBO as our primary baseline because it shares an ensemble-based structure with our method, allowing for a direct comparison of the stochastic interaction mechanism.
    \item \textbf{Stochastic Gradient Descent (SGD):} Included as a classical baseline for local-trajectory-based search to contrast with the global population-based approach.
\end{itemize}

\subsubsection{Benchmark Function: The Ackley Landscape}
To validate the performance of our framework across different scales, we utilize the Ackley ($A$) function. This function is a standard benchmark for global optimization due to its nearly flat outer region and a large number of local minima that intensify toward the global optimum:
\begin{equation}
    A(x) = -20 \exp\left(-0.2 \sqrt{\frac{1}{d}\sum_{i=1}^{d} x_i^2}\right)
    - \exp\left(\frac{1}{d}\sum_{i=1}^{d} \cos(2\pi x_i)\right) + 20 + e,
\end{equation}
where $x \in \mathbb{R}^d$. The function has a unique global minimum at $x^\ast=0$ with $A(x^\ast)=0$. We evaluate the algorithms for $d \in \{1, 2, 10, 30\}$.

\subsubsection{Initialization Regimes}
To test the geometric selection of the search process, we consider two initialization scenarios

\begin{itemize}
    \item \textbf{Uniform Initialization:} Individuals are drawn from $\mathcal{U}(-2,2)^d$. In this regime, the global optimum is encapsulated within the initial support of the population.
    \item \textbf{Shifted Normal Initialization:} The population is initialized from a Normal distribution $\mathcal{N}(\mu,\sigma_\text{init}^2 I)$ with $\mu=2.0$ and $\sigma_\text{init}=0.5$. This creates a ``blind'' search scenario where the optimum is outside the initial support, requiring effective global transport.
\end{itemize}

\subsubsection{Hyperparameters and Metrics}

For a rigorous comparison, parameters were aligned across all frameworks to ensure equivalent selection and diffusion strengths. The population-based methods, CES and CBO, share a population size of $M=20,000$ with $a=\lambda_{\text{CBO}}=40$, $c=\sigma_{\text{CBO}}=0.04$, and a final time $T_f=1.0$.

The temporal discretization was defined as follows
\begin{itemize}
    \item \textbf{CES:} The number of iterations is set to $N_t = \lceil T_f \cdot M^{\alpha} \rceil$ with $\alpha=0.5$, resulting in $141$ steps.
    \item \textbf{CBO:} To ensure parity, the CBO baseline was restricted to the same budget of $N_t = 141$ iterations.
    \item \textbf{SGD:} As a standard baseline, SGD was granted $20,000$ iterations with a learning rate $\nu=10^{-2}$ to compensate the lack of ensemble information.
\end{itemize}

The performance is measured using the Best-of-Ensemble $L_2$ error, defined as the distance between the best-performing individual at the final time $T_f$ and the global optimum
\begin{equation}
    \text{Error}_{L_2}=\| \arg \min_{i=1,\dots,M} f(x_i(T_f)) - x^\ast \|_2.
\end{equation}
We also report the Success Rate, defined as the percentage of runs achieving $\text{Error}_{L_2}<10^{-2}$.

\begin{remark}{Implementation Details for Minimization.}
To adapt the theoretical framework to the minimization of the loss function $A(x)$, we define the selection weights as $w_M(x) := e^{-a_M A(x)}$. This exponential mapping ensures that the selection probability remains strictly positive and well-defined across the entire domain $\mathbb{R}^d$. For large populations $M$, this choice is asymptotically consistent with the linear fitness $\sigma = -A$ used in our derivation, as $e^{-a_M A} \approx 1 + a_M \sigma + o(a_M)$. Also, in the unbounded domain $\mathbb{R}^d$, the mutation step is implemented via a multivariate Normal distribution $\mathcal{N}(y, 2tI_d)$. 
\end{remark}

\subsection{Analysis of Numerical Results}

The comparative performance of CES, CBO, and SGD across different dimensions and initialization regimes is summarized in Table \ref{tab:final_results}. We evaluate the performance across three key aspects: convergence precision, mass transport under shifted initializations, and high-dimensional scalability.

\begin{table}[tb]
\centering
\caption{Benchmark Performance: Mean $L_2$ Error $\pm$ SD and Success Rate (SR).}
\label{tab:final_results}
\resizebox{\textwidth}{!}{%
\begin{tabular}{ll cc cc cc}
\toprule
\textbf{Scenario} & \textbf{Dim} & \multicolumn{2}{c}{\textbf{CES (Ours)}} & \multicolumn{2}{c}{\textbf{CBO}} & \multicolumn{2}{c}{\textbf{SGD}} \\
\cmidrule(lr){3-4} \cmidrule(lr){5-6} \cmidrule(lr){7-8}
& & \textit{Error ($\pm$ SD)} & \textit{SR} & \textit{Error ($\pm$ SD)} & \textit{SR} & \textit{Error ($\pm$ SD)} & \textit{SR} \\
\midrule
\multirow{4}{*}{\textbf{Uniform}} & $d=1$ & \textbf{\errCESUniOne} & \textbf{\srCESUniOne} & \errCBOUniOne & \srCBOUniOne & \errSGDUniOne & \srSGDUniOne \\
 & $d=2$ & \errCESUniTwo & \srCESUniTwo & \textbf{\errCBOUniTwo} & \textbf{\srCBOUniTwo} & \errSGDUniTwo & \srSGDUniTwo \\
 & $d=10$ & \errCESUniTen & \srCESUniTen & \textbf{\errCBOUniTen} & \textbf{\srCBOUniTen} & -- & -- \\
 & $d=30$ & \errCESUniThirty & \srCESUniThirty & \textbf{\errCBOUniThirty} & \textbf{\srCBOUniThirty} & -- & -- \\
\midrule
\multirow{4}{*}{\textbf{Shifted}} & $d=1$ & \textbf{\errCESShiOne} & \textbf{\srCESShiOne} & \errCBOShiOne & \srCBOShiOne & \errSGDShiOne & \srSGDShiOne \\
 & $d=2$ & \textbf{\errCESShiTwo} & \textbf{\srCESShiTwo} & \errCBOShiTwo & \srCBOShiTwo & \errSGDShiTwo & \srSGDShiTwo \\
 & $d=10$ & \textbf{\errCESShiTen} & \textbf{\srCESShiTen} & \errCBOShiTen & \srCBOShiTen & -- & -- \\
 & $d=30$ & \textbf{\errCESShiThirty} & \textbf{\srCESShiThirty} & \errCBOShiThirty & \srCBOShiThirty & -- & -- \\
\bottomrule
\end{tabular}%
}
\end{table}

\subsubsection{Precision and Consensus Bias (Uniform Case)}
In the Uniform regime, where the global optimum is within the initial population support, both ensemble methods perform competitively in low dimensions ($d=1,2$).
\begin{itemize}
    \item \textbf{CES precision:} Our method achieves a $\srCESUniOne$ SR with errors as low as \errCESUniOne, demonstrating high-exploitation capabilities.
    \item \textbf{CBO Efficiency:} While CBO shows slightly better mean errors in $d=10$ and $d=30$ for the uniform case, its SR drops nearly to zero. This confirms that while CBO quickly approaches the optimum area, it often lacks the final exploratory pressure to fall below the success threshold $\epsilon=10^{-2}$.
    \item \textbf{SGD Stagnation:} Despite having $20.000$ iterations, SGD consistently fails ($\srSGDUniOne$ SR) in all dimensions. The near-zero standard deviation (\errSGDUniTwo) indicates that SGD is deterministically trapped in a local minima across all runs, highlighting the limitations of local-trajectory searchers in the rugged Ackley landscape.
\end{itemize}

\subsubsection{Global Mass Transport}
The Shifted scenario reveals the most significant advantage of the CES framework. When the population is initialized away from the optimum, the algorithm must perform effective mass transport across multiple local basins.

\begin{figure}[tb]
\centering
\hfill
\begin{subfigure}{0.48\textwidth}
\centering
\includegraphics[width=\textwidth]{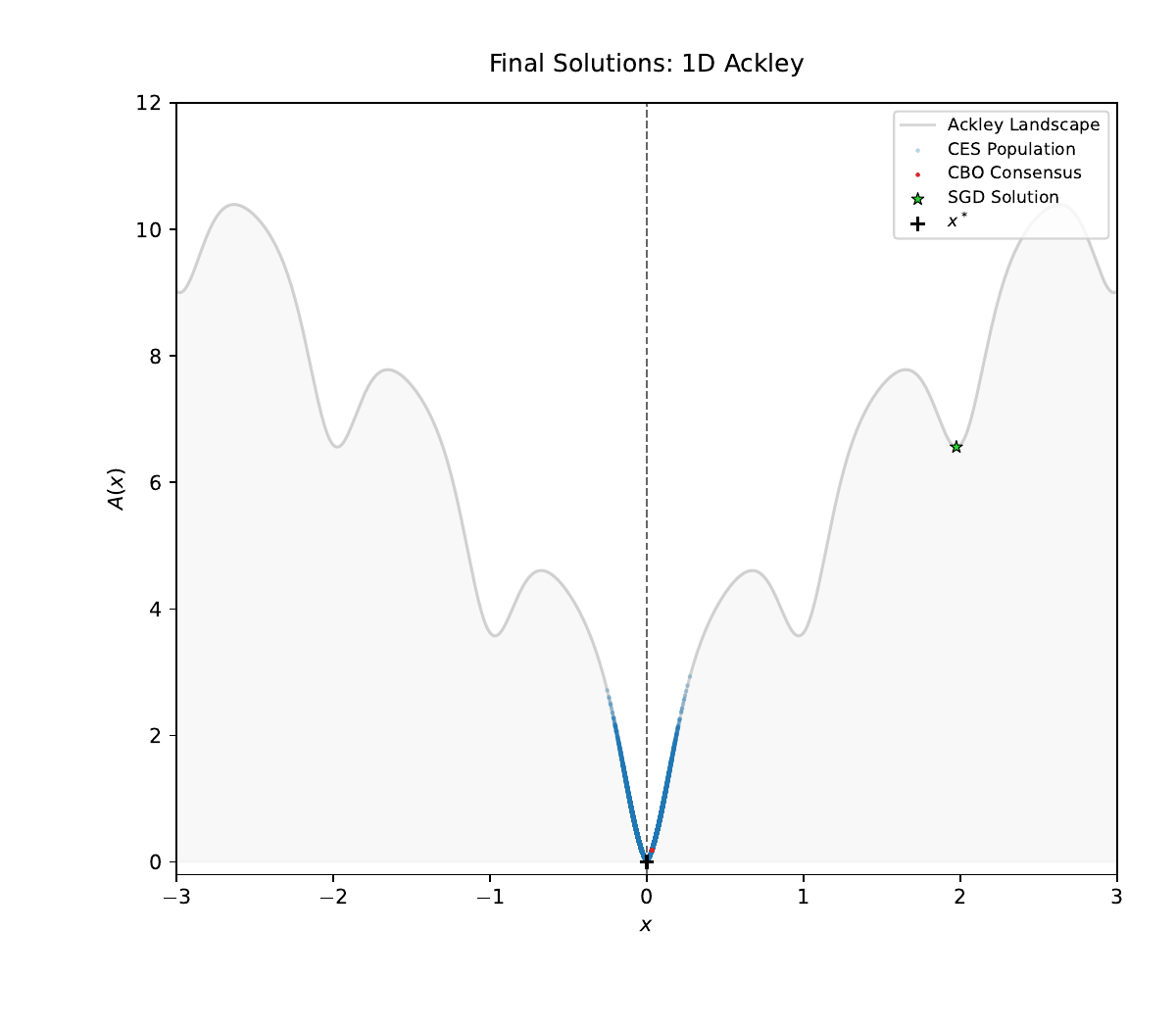}
\caption{Shifted case: $d=1$.}\label{fig:transport_dynamics__a}
\end{subfigure}
\hfill
\begin{subfigure}{0.48\textwidth}
\centering
\includegraphics[width=\textwidth]{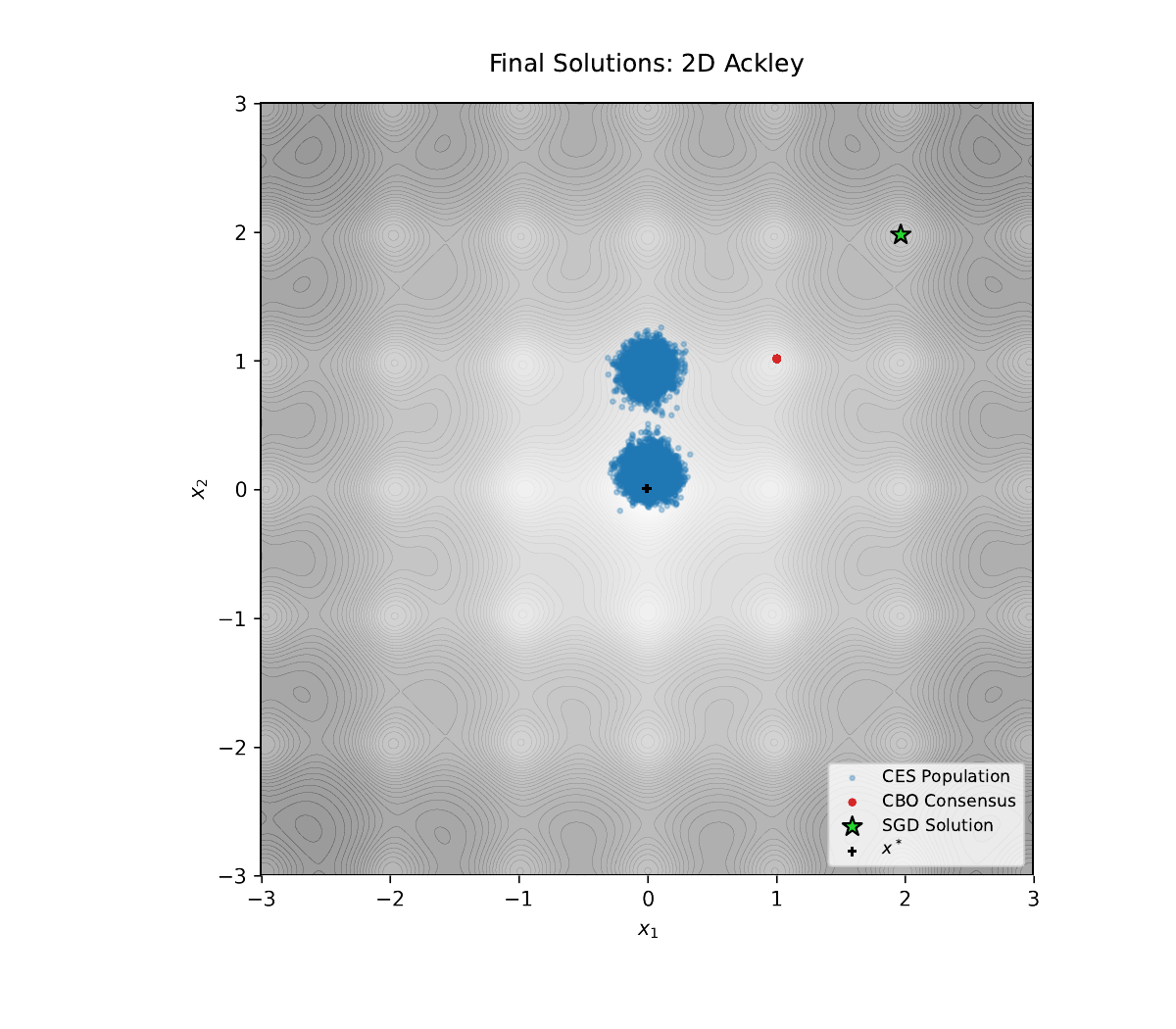}
\caption{Shifted case: $d=2$.}\label{fig:transport_dynamics__b}
\end{subfigure}
\hfill
\caption{Population mass transport in the Shifted scenario ($d=1$ and $d=2$). While CES successfully migrates its mass toward the global optimum at $x^\ast=0$ by overcoming multiple local basins, the baselines exhibit stagnation: in $d=1 $, CBO collapses near the optimum, whereas in $d=2$ it remains trapped far from the target. In all cases, SGD stays localized in the nearest local minima, failing to perform any global transport.}
\label{fig:transport_dynamics}
\end{figure}

\begin{itemize}
    \item \textbf{Superiority in $d=1$ and $d=2$:} As illustrated in our population distribution (Figure \ref{fig:transport_dynamics}). CES successfully migrates its mass toward the global optimum, maintaining a $\srCESShiOne$ SR in $d=1$ and $\srCESShiTwo$ SR in $d=2$. In contrast, CBO fails to collapse close enough to the optimum in $d=1$ ($\srCBOShiOne$), and in $d=2$ stagnates in local basins closer to its initial mean.
    \item \textbf{Failure of Baseline Consensus:} The CBO's failure in the shifted case in $d=2$ (CBO \srCBOShiTwo vs. CES \srCESShiTwo) suggests that the consensus mechanism is highly sensitive to the initial ``gravitational'' center of the population.
\end{itemize}

\subsubsection{Scalability and Convergence Rates}

In high-dimensional settings ($d=30$), the curse of dimensionality affects both ensemble methods, yet the relative performance of CES remains superior in the shifted regime.

\begin{figure}[tb]
    \centering
    \includegraphics[width=0.92\textwidth]{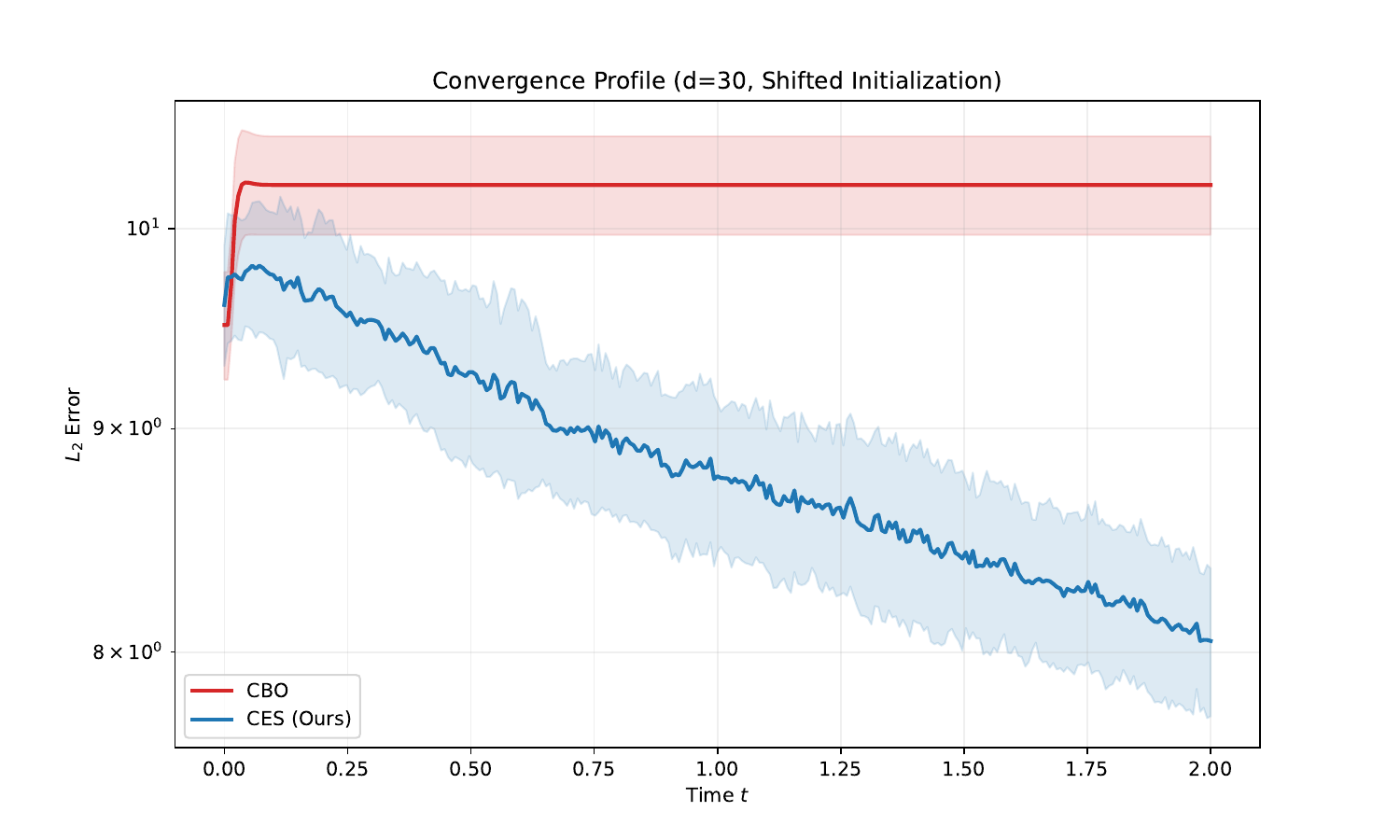}
    \caption{Convergence rates for the Shifted $d=30$ scenario. CBO reach a premature plateau, while CES maintains a consistent descent.}
    \label{fig:convergence_30d}
\end{figure}

\begin{itemize}
    \item \textbf{Robustness} In the Shifted $d=30$ case, CES achieves a mean error of \errCESShiThirty, lower than CBO's \errCBOShiThirty.
    \item \textbf{Convergence Dynamics:} Our convergence rate (Figure \ref{fig:convergence_30d}) show that while CBO reaches a plateau early (stagnation), CES continues to refine its position. This behavior is consistent with the geometric selection property discussed in the 1D theoretical analysis: the mutational pressure in CES prevents premature collapse, allowing the population to navigate the high-dimensional landscape more effectively.
\end{itemize}

\section{Discussion}
%\todo[inline]{Reviewer: The ``Geometric Selection / survival of the flattest'' framing will interest the deep-learning optimisation community. Please broaden citations here beyond [13] to include the flat-minima literature: Hochreiter \& Schmidhuber (1997, \emph{Flat Minima}), Keskar et al.\ (2017, ICLR) on large-batch training and sharp minima, Dinh et al.\ (2017) on sharp minima that generalise, and Foret et al.\ (2021, SAM). The spectral-concentration perspective offers a genuinely new lens on this debate and deserves to be positioned explicitly.}
%\todo[inline]{Reviewer: In §5.2 the word ``tunneling'' is used loosely. Either make the analogy precise (via exponential decay of the semiclassical ground state across a potential barrier, $\sim e^{-d(x,x^*)/h}$) or soften the wording to ``mass transport through low-fitness regions'' to avoid overstating the quantum-mechanical parallel.}

The results presented in this work bridge the gap between biological evolution models and global optimization, providing a rigorous semiclassical framework to understand population dynamics.

\subsection{The Semiclassical Convergence: From Individuals to Eigenvalues}

Beyond establishing a convergence framework, the Schrödinger-type representation reveals fundamental structural properties of the optimization dynamics. The analysis relies on the following mathematical hierarchy

\begin{itemize}
    \item \textbf{Macroscopic Limit:} We bridge the individual-based dynamics of the CES to a deterministic mean-field limit. This transition, rooted in the probabilistic limits of evolutionary ecology \cite{fournier2004microscopic,champagnat2006unifying}, yields a replicator-mutator equation governing the population density. This establishes a Wright-Fisher evolutionary counterpart to the McKean-Vlasov processes standardly used in CBO \cite{Pinnau2016ConsensusMeanField,carrillo2018analytical}.
    \item \textbf{Semiclassical Scaling:} We map the search for a global optimum to the spectral analysis of a Schrödinger-type operator. By framing the fitness landscape as a potential well, this connection leverages tools from diffusion-selection systems and the study of metastable states.
    \item \textbf{Spectral Concentration:} In the semiclassical limit, global convergence is governed by the principal eigenfunction of the system. The population mass concentrates on the ground state of the operator, providing an optimization counterpart to the spectral analysis of the replicator-mutator equations.
\end{itemize}

Consequently, this framework formalizes the mechanism of \emph{geometric selection}, offering a rigorous mathematical justification for the survival of the flattest phenomenon. While empirically observed in biology \cite{alfaro2019evolutionary} and deep learning \cite{hochreiter1997flat}, our spectral analysis demonstrates that CES inherently favors wider, more robust regions of the landscape --- those with higher spectral weight---thereby avoiding narrow local traps. Unlike traditional multi-agent methods that may suffer from premature variance collapse into sub-optimal basins due to a lack of geometric awareness \cite{fornasier2026}, the global convergence of the CES is characterized here as an intrinsic property of the operator's spectrum.

\subsection{CES vs. CBO: Mechanism of Exploration}

The stark difference in performance under Shifted Initialization (Table \ref{tab:final_results}) reveals a fundamental distinction in their stochastic mechanism of exploration
\begin{itemize}
    \item \textbf{Consensus Collapse:} CBO relies on a weighted average (consensus) which, under shifted conditions, often falls into a consensus bias \cite{Pinnau2016ConsensusMeanField,carrillo2018analytical}. If the initial mass is far from the optimum, the multiplicative noise in CBO may lead to a premature collapse of the variance toward sub-optimal states \cite{fornasier2026}.
    \item \textbf{Global Mass Transport:} CES utilizes additive mutation (diffusion) coupled with replicator dynamics. This allows the population to behave like a migrating wave front. Even if the optimum is outside the initial support, the mass transport through low-fitness regions effect provided by the mutation-selection balance facilitates global transport, as seen in Figure \ref{fig:transport_dynamics}.
\end{itemize}

Recent advancements in consensus-based methods, such as $\delta$-CBO and the Consensus Freezing schemes proposed by \cite{fornasier2026}, attempt to mitigate this premature collapse by introducing non-vanishing diffusion to prevent the noise term from vanishing too early. While these \emph{consensus hopping} variants -- which can be interpreted as a form of $(1, \lambda)$-Evolution Strategy -- improve robustness, they rely on sophisticated numerical patches like rescaled time-steps and freezing intervals to maintain stability \cite{fornasier2026}.

Our proposed CES framework addresses the same challenge intrinsically through the replicator-mutator dynamics. Furthermore, while numerical benchmarks for recent CBO variants often demonstrate success under a convergence threshold of $\epsilon=0.1$ \cite{fornasier2026}, our results suggest that the CES framework remains robust under the significantly stricter thresholds (e.g., $10^{-2}$) typically required for high-precision global optimization \cite{Hansen_2020}. This indicates that the replicator-mutator dynamics may offer a more direct and precise path toward exact localization of the global optimum in high-dimensional settings without the need for additional numerical heuristics.

\section{Conclusions and Future Work}

We have framed the CES and established its mean-field limit as a Schrödinger-type equation. Our findings demonstrate that the semiclassical concentration of the stationary state provides a formal path to prove global convergence, acting as a geometric filter that prioritizes robust optima (``survival of the flattest''). This allows CES to maintain superior mass transport capabilities compared to CBO and SGD, especially in high-dimensional ($d=30$) rugged landscapes where traditional methods suffer from premature consensus collapse.

Natural extensions of this work include generalizing the mean-field derivation to $d>1$ and exploring the CES as a mesh-free solver for high-dimensional PDEs, where population-based concentration can mitigate the curse of dimensionality. Furthermore, we aim to incorporate more complex evolutionary operators, such as crossover (recombination), into the mean-field framework to study how multi-parent interactions affect the spectral concentration properties. Finally, a more extensive benchmark analysis involving non-separable and ill-conditioned functions will be conducted to further validate the robustness of the Schrödinger-based approach in large-scale industrial applications.

The goal of this contribution was report our initial results. However, we are aware the further experimentation is important in order to establish CES ranking with respect to the state of the art. We will report those results online (see below) and complement the paper with those results.
\begin{credits}

\subsubsection{\ackname} Work funded by the Franco-Chilean Binational Center of Artificial Intelligence, ANID Strengthening R\&D capabilities Program CTI230007 Inria Chile, and Inria Challenge OcéanIA (desc. num 14500).

\subsubsection{\discintname}
The authors declare no competing interests.

\subsubsection{Code Availability}
The code supporting our experiments and the results of the paper are available online at \url{https://github.com/Inria-Chile/ces-global-convergence}.
\end{credits}

\bibliographystyle{splncs04}
\bibliography{convergence-eas}

\begin{thebibliography}{10}
\providecommand{\url}[1]{\texttt{#1}}
\providecommand{\urlprefix}{URL }
\providecommand{\doi}[1]{https://doi.org/#1}

\bibitem{alfaro2019evolutionary}
Alfaro, M., Veruete, M.: Evolutionary branching via replicator-mutator
  equations. Journal of Dynamics and Differential Equations  \textbf{31},
  2029--2052 (2019). \doi{10.1007/s10884-018-9692-9}

\bibitem{cantrell2003spatial}
Cantrell, R.S., Cosner, C.: Spatial Ecology via Reaction-Diffusion Equations.
  Series in Mathematical and Computational Biology, John Wiley \& Sons,
  Chichester, UK (2003)

\bibitem{carrillo2018analytical}
Carrillo, J.A., Choi, Y.P., Totzeck, C., Tse, O.: An analytical framework for
  consensus-based global optimization method. Mathematical Models and Methods
  in Applied Sciences  \textbf{28}(06),  1037--1066 (2018).
  \doi{10.1142/S0218202518500276}

\bibitem{champagnat2006unifying}
Champagnat, N., Ferrière, R., Méléard, S.: Unifying evolutionary dynamics:
  From individual stochastic processes to macroscopic models. Theoretical
  Population Biology  \textbf{69}(3),  297--321 (2006).
  \doi{10.1016/j.tpb.2005.10.004}

\bibitem{Coville2013Convergence}
Coville, J.: Convergence to equilibrium for positive solutions of some
  mutation-selection model (2013), \url{https://arxiv.org/abs/1308.6471}

\bibitem{fontbona2015nonlocal}
Fontbona, J., M{\'e}l{\'e}ard, S.: Non local lotka-volterra system with
  cross-diffusion in an heterogeneous medium. Journal of Mathematical Biology
  \textbf{70}(4),  829--854 (2015). \doi{10.1007/s00285-014-0781-z}

\bibitem{fornasier2026}
Fornasier, M., Huang, H., Klemenc, J., Malaspina, G.: From consensus-based
  optimization to evolution strategies: Proof of global convergence. arXiv
  preprint  (2026), \url{https://arxiv.org/abs/2602.11677}

\bibitem{fournier2004microscopic}
Fournier, N., Méléard, S.: A microscopic probabilistic description of a
  locally regulated population and macroscopic approximations. The Annals of
  Applied Probability  \textbf{14}(4) (2004). \doi{10.1214/105051604000000882}

\bibitem{gendreau2013handbook}
Gendreau, M., Potvin, J.Y.: Handbook of Metaheuristics. Springer (2013)

\bibitem{Hansen_2020}
Hansen, N., Auger, A., Ros, R., Mersmann, O., Tušar, T., Brockhoff, D.: Coco:
  a platform for comparing continuous optimizers in a black-box setting.
  Optimization Methods and Software  \textbf{36}(1),  114–144 (2020).
  \doi{10.1080/10556788.2020.1808977}

\bibitem{hansen2001completely}
Hansen, N., Ostermeier, A.: Completely derandomized self-adaptation in
  evolution strategies. Evolutionary Computation  \textbf{9}(2),  159--195
  (2001). \doi{10.1162/106365601750190398}

\bibitem{herty2024multiscale}
Herty, M., Huang, Y., Kalise, D., Kouhkouh, H.: A multiscale consensus-based
  algorithm for multilevel optimization. Mathematical Models and Methods in
  Applied Sciences  \textbf{35}(10),  2207–2243 (Jun 2025).
  \doi{10.1142/s021820252550037x}

\bibitem{hochreiter1997flat}
Hochreiter, S., Schmidhuber, J.: Flat minima. Neural Computation
  \textbf{9}(1),  1--42 (01 1997). \doi{10.1162/neco.1997.9.1.1}

\bibitem{meleard2016modeles}
M{\'e}l{\'e}ard, S.: Mod{\`e}les al{\'e}atoires en Ecologie et Evolution.
  Springer Berlin, Heidelberg (2016)

\bibitem{10.1098/rsfs.2013.0054}
Morozov, A.: Modelling biological evolution: recent progress, current
  challenges and future direction. Interface Focus  \textbf{3}(6),  20130054
  (12 2013). \doi{10.1098/rsfs.2013.0054},
  \url{https://doi.org/10.1098/rsfs.2013.0054}

\bibitem{pathiraja2024mathematicaldescriptioncontinuoustime}
Pathiraja, S., Wacker, P.: Mathematical description of continuous time and
  space replicator-mutator equations for quadratic fitness landscapes (2024),
  \url{https://arxiv.org/abs/2412.08178}

\bibitem{Pinnau2016ConsensusMeanField}
Pinnau, R., Totzeck, C., Tse, O., Martin, S.: A consensus-based model for
  global optimization and its mean-field limit. Mathematical Models and Methods
  in Applied Sciences  \textbf{27}(01),  183–204 (2017).
  \doi{10.1142/s0218202517400061}

\bibitem{roith2025consensus}
Roith, T., Bungert, L., Wacker, P.: Consensus-based optimization for closed-box
  adversarial attacks and a connection to evolution strategies (2025),
  \url{https://arxiv.org/abs/2506.24048}

\bibitem{wakano2017derivation}
Wakano, J.Y., Funaki, T., Yokoyama, S.: Derivation of replicator--mutator
  equations from a model in population genetics. Japan Journal of Industrial
  and Applied Mathematics  \textbf{34},  473--488 (2017).
  \doi{10.1007/s13160-017-0249-9}

\bibitem{refId0}
{Weiss, Pierre}: L'hypoth{\`e}se du champ mol{\'e}culaire et la
  propri{\'e}t{\'e} ferromagn{\'e}tique. J. Phys. Theor. Appl.  \textbf{6}(1),
  661--690 (1907). \doi{10.1051/jphystap:019070060066100},
  \url{https://doi.org/10.1051/jphystap:019070060066100}

\bibitem{williams2017mode}
Williams, G., Aldrich, A., Theodorou, E.A.: Model predictive path integral
  control: From theory to parallel computation. Journal of Guidance, Control,
  and Dynamics  \textbf{40}(2),  344--357 (2017). \doi{10.2514/1.G001921}

\bibitem{Zworski2012Semiclassical}
Zworski, M.: Semiclassical Analysis, Graduate Studies in Mathematics, vol.~138.
  American Mathematical Society, Providence, RI (2012)

\end{thebibliography}
\end{document}